\documentclass[runningheads]{llncs}
\usepackage{silence}
\usepackage{amsmath}
\usepackage{amsfonts} 
\usepackage{float}
\usepackage[T1]{fontenc}
\usepackage{placeins}

\usepackage{graphicx}
\usepackage{enumitem}
\usepackage{url}
\DeclareUnicodeCharacter{202F}{\,}
\begin{document}
\title{Biological Processing Units: Leveraging an Insect  Connectome to Pioneer Biofidelic Neural Architectures}
\titlerunning{Biological Processing Units}
%
\author{Siyu Yu\inst{1,2,\thanks{Equal contribution}}\orcidID{0009-0001-1459-2567} \and Zihan Qin\inst{1,\footnotemark[1]}\orcidID{0009-0000-8261-5786} \and\\[-1ex]
Tingshan Liu\inst{1,\footnotemark[1]}\orcidID{0000-0003-0232-2495} \and
Beiya Xu\inst{1}\orcidID{0009-0002-1462-3850}
\and\\[0.15ex] R. Jacob Vogelstein\inst{3}\orcidID{0000-0001-8159-9577}
\and Jason Brown\inst{2}
\and 
\\[0.15ex] Joshua T. Vogelstein\inst{1,2}\orcidID{0000-0003-2487-6237} }
%
\authorrunning{Yu \textit{et al.}}
%
\institute{Johns Hopkins University, Baltimore MD 21218, USA 
\and Pomegranate Intelligence, Baltimore MD 21218, USA
\and Catalio Capital Management, New York NY 10010, USA\\
\email{\{syu80, zqin16, tliu68, bxu41, jovo\}@jhu.edu}\\\email{jason@pomintel.com}\quad\email{rjv@cataliocapital.com}}
%
\maketitle              
%
\begin{abstract}
The complete connectome of the \textit{Drosophila} larva brain offers a unique opportunity to investigate whether biologically evolved circuits can support artificial intelligence. We convert this wiring diagram into a Biological Processing Unit (BPU)—a fixed recurrent network derived directly from synaptic connectivity. Despite its modest size (3{,}000 neurons and 65{,}000 weights between them), 
the unmodified BPU achieves 98\% accuracy on MNIST and 58\% on CIFAR-10, surpassing size-matched MLPs. Scaling the BPU via structured connectome expansions further improves CIFAR-10 performance, while modality-specific ablations reveal the uneven contributions of different sensory subsystems.
On the ChessBench dataset, a lightweight GNN-BPU model trained on only 10{,}000 games achieves 60\% move accuracy, nearly 10x better than any size transformer.
Moreover, CNN-BPU models with $\sim$2M parameters outperform parameter-matched Transformers, and with a depth-6 minimax search at inference, reach 91.7\% accuracy, exceeding even a 9M-parameter Transformer baseline.
These results demonstrate the potential of biofidelic neural architectures to support complex cognitive tasks and motivate scaling to larger and more intelligent connectomes in future work.


\end{abstract}

\section{Introduction}
The recent completion of the entire \textit{Drosophila} larval connectome, comprising approximately 3000 neurons and 65,000 weights between them, provides a rare opportunity to examine a fully natural-optimized neural circuit \cite{winding2023connectome}. In contrast to large-scale artificial models that often require extensive computation and tuning, biological systems like \textit{Drosophila} achieve complex behaviors with minimal resources. This suggests that a complete biological connectome may serve as a biological lottery ticket \cite{frankle2019lotterytickethypothesisfinding,pmlr-v119-hasani20a}: a compact, evolutionarily selected circuit capable of supporting a broad range of cognitive functions.

Previous studies \cite{lappalainen2024connectome,wang2023neuroinspired,liang2021fruitflylearnword} have leveraged partial connectome structures from adult \textit{Drosophila} \cite{zheng2018complete} to guide neural network design, demonstrating the promise of biologically inspired architectures. However, such approaches may miss critical dynamics and functional motifs present only in the complete connectome. With the full larval connectome now available, we hypothesize that a fully intact biological neural circuit can inform the design of efficient and generalizable artificial systems, as it embodies solutions to many of the same computational challenges neural networks aim to address. To test this, we directly employ the complete connectome without altering its structure or synaptic weights, assessing whether it can support diverse cognitive tasks without task-specific adaptation.

Here we directly leverage the complete \textit{Drosophila} larval connectome to develop Biological Processing Units (BPUs). We evaluate BPU on two categories of tasks: sensory processing (MNIST, CIFAR-10) \cite{lappalainen2024connectome} and decision-making (chess puzzles) \cite{ruoss2024amortized}. These tasks are chosen to reflect fundamental cognitive functions—perception, memory, and planning—that are intrinsic to both artificial and biological agents. By including peripheral sensors alongside the central BPU circuit, we test whether the BPU can support generalized cognition under realistic biological constraints. Finally, to understand how far
this advantage can scale, we introduce a directed, signed
degree–corrected Stochastic Block Model (DCSBM) that lets us expand the larval connectome up to 5× while faithfully preserving
its block-level wiring statistics and synaptic polarity.

The BPU achieves competitive performance across all tasks, matching or surpassing baseline models with similar parameter counts. These results support the idea that intact biological connectomes can serve as effective, reusable substrates for intelligent computation.

\section{Methods}
\subsection{BPU architecture}
We embed the entire larval \textit{Drosophila} connectome as a fixed-weight recurrent core. The synaptic weights are directly taken from the connectome and remain unchanged during training. Only the input and output projections are optimized via gradient descent. Over a fixed number of unrolled steps, activity propagates through the reservoir. The trainable projections map inputs to internal dynamics and decode them into task-relevant outputs.

\begin{figure}[!ht]
    \centering
    \includegraphics[width=\linewidth]{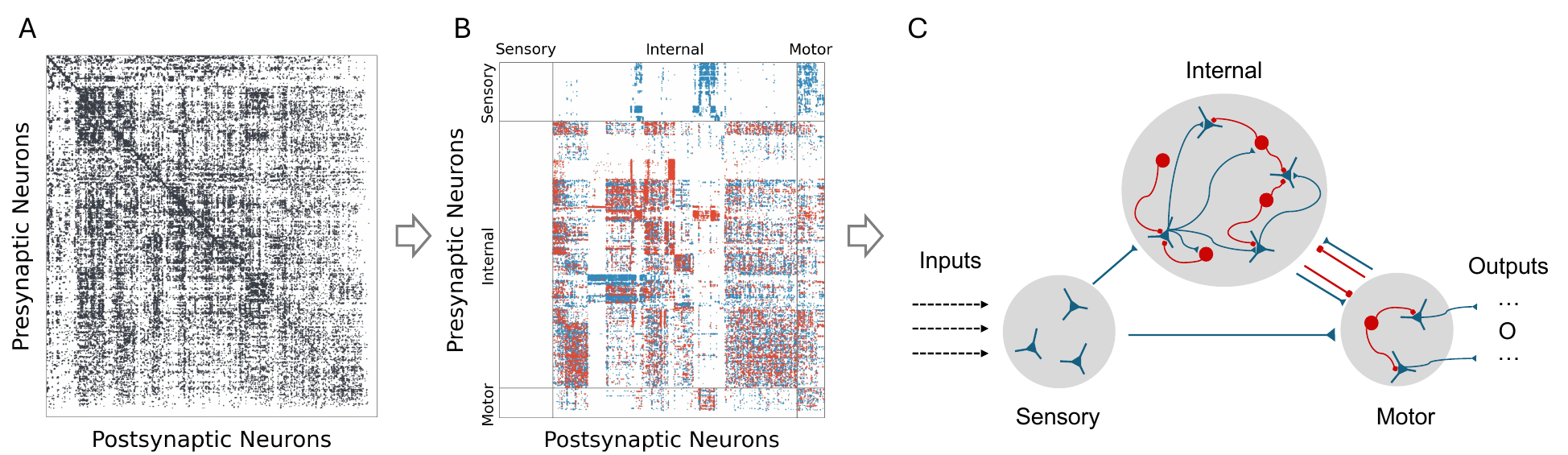}
    \caption{\textbf{Biological Processing Unit (BPU) architecture based on the larval \textit{Drosophila} connectome.} 
    \textbf{(A)} Raw axon-to-dendrite adjacency matrix representing synaptic connectivity. 
    \textbf{(B)} Signed connectivity matrix after applying neurotransmitter-derived polarities and partitioning neurons into sensory, internal, and output pools. 
    \textbf{(C)} Schematic of the BPU: inputs project to sensory neurons, activity propagates through the fixed recurrent core, and outputs are read from designated output neurons. 
    In (B) and (C), blue denotes excitatory connections and red denotes inhibitory ones.}
    \label{fig:architecture}
\end{figure}
 As illustrated in Figure~\ref{fig:architecture}, the BPU utilizes the axon-to-dendrite connectivity adjacency matrix derived from electron microscopy reconstructions~\cite{lappalainen2024connectome}. We assign directional polarity (excitatory or inhibitory) to each connection by multiplying synaptic counts with neurotransmitter-based annotations~\cite{wang2024we}. The neurons are partitioned into three functionally distinct pools based on anatomical annotations, while retaining all neurons within the recurrent computational core:
\begin{itemize}[leftmargin=1.5em, label=\scriptsize$\bullet$]
  \item Sensory ($N = 430$): neurons responsible for encoding external stimuli;
  \item Output ($N = 218$): descending neurons projecting to motor circuits (DN-SEZ) and ring gland neurons (RGN) targeting neuroendocrine structures;
  \item Internal neurons ($N = 2304$): all other neurons
\end{itemize}

The BPU’s recurrent dynamics evolve according to:
\begin{equation}
\begin{aligned}
S(t{+}1) &= f\big(W_{\text{ss}} S(t) + W_{\text{rs}} I(t) + W_{\text{os}} O(t) + E(t)\big) \\
I(t{+}1) &= f\big(W_{\text{sr}} S(t) + W_{\text{rr}} I(t) + W_{\text{or}} O(t)\big) \\
O(t{+}1) &= f\big(W_{\text{so}} S(t) + W_{\text{ro}} I(t) + W_{\text{oo}} O(t)\big)
\end{aligned}
\label{eq:bpu_dynamics}
\end{equation}

Here, $f(\cdot)$ denotes a nonlinear activation function (typically ReLU), and $\mathrm{W}_{xy}$ represent fixed, connectome-derived synaptic weight matrices. To preserve biological plausibility, we constrain the temporal depth of recurrent processing (i.e., the number of time steps $T$) to match the characteristic synaptic propagation path length observed in the \textit{Drosophila} sensory pathways.

\subsection{Connectome expansion via a directed, signed DCSBM}
To explore how scale influences performance, we stochastically enlarge the larval connectome using a directed, signed degree–corrected stochastic block model (DCSBM) \cite{karrer2011stochastic,gao2018community}.
Let $W\in\mathbb{R}^{N_0\times N_0}$ denote the signed adjacency of the empirical core and $z_i\!\in\!\{0,1,2\}$ its sensory/internal/output labels.  We fit a DCSBM with separate out- and in-strengths $\theta^{\mathrm{out}}_i,\theta^{\mathrm{in}}_i$, block–pair weight densities $\omega_{gh}$ (Eq.~\ref{eq:omega}) and sign probabilities $p_{gh}$ (Eq.~\ref{eq:pgh}).

\begin{align}
    \omega_{gh} &=
      \frac{\sum_{i\in g}\sum_{j\in h}|W_{ij}|}
           {\bigl(\sum_{i\in g}\theta^{\mathrm{out}}_i\bigr)
            \bigl(\sum_{j\in h}\theta^{\mathrm{in}}_j\bigr)},
    \label{eq:omega}\\
    p_{gh} &=
      \frac{\#\{\,W_{ij}>0\,|\,z_i=g,z_j=h\}}
           {\#\{\,W_{ij}\neq0\,|\,z_i=g,z_j=h\}}.
    \label{eq:pgh}
\end{align}
To obtain a target size $N=F\,N_0$ (expansion factor $F\!\in\!\{1,2,...,5\}$) we first bootstrap paired in- and out-degrees $(\theta^{\text{out}},\theta^{\text{in}})$ from core neurons within the same block and then rescale the two vectors so that $\sum\theta^{\text{out}}=\sum\theta^{\text{in}}$.
Then we draw block labels by the empirical proportion $$\Pr(z=k)=|z_i=k|/N_0$$ For every ordered node pair $(u,v)$, we sample a Poisson edge count $$\Lambda_{uv}\!\sim\!\mathrm{Poisson}(\theta^{\mathrm{out}}_u \theta^{\mathrm{in}}_v\omega_{z_u z_v})$$ and assign its polarity by a Bernoulli draw with probability $p_{z_u z_v}$.
 Finally, the original $N_0\times N_0$ sub-matrix is restored exactly so all experiments remain anchored in the real connectome.

\subsection{Baseline for image classification}
To adapt standard image datasets for the BPU architecture, we flatten each image and project it into the BPU’s sensory neuron subspace. The MNIST input (784 dimensions) and the CIFAR-10 input (3,072 dimensions) are linearly mapped to match the size of the sensory input. The resulting vector serves as the external input $\mathrm{E}(t)$ in $t = 0$, as defined in Equation~\ref{eq:bpu_dynamics}.

To isolate architectural effects, we use a two‑hidden‑layer MLP in which only the input-to-first-hidden and second-hidden-to-output mappings are trainable, together matching exactly the BPU projection parameter count, while the intermediate hidden-to-hidden transform remains a fixed untrained random projection. Activations (ReLU) and optimization mirror BPU settings.

\subsection{BPU for Chess Puzzle Solving}
For the chess task, we evaluate the BPU using \textit{puzzle accuracy}, the percentage of puzzles where the predicted move sequence exactly matches the full ground-truth solution. Each puzzle is drawn from a curated Lichess dataset~\cite{carlini2023playing}, annotated with Elo difficulty ratings ranging from 399 to 2867 and complete solution sequences.

We use the ChessBench dataset~\cite{lappalainen2024connectome}, which provides $10 \times 10^6$ board positions sourced from Lichess.org games. Each state is encoded using the Forsyth-Edwards Notation (FEN)~\cite{edwards1994standard}, and annotated using Stockfish 16 under 50 ms per board constraints. The state value labels reflect the estimated win probabilities between 0\% and 100\%. To convert FEN strings into fixed-size neural inputs for the BPU, we implement two encoding pipelines:

\textbf{GNN-based encoder.}  
We represent each FEN position as a 65-node directed graph: 
the 64 board squares plus a central “hub”. 
Square nodes carry a 12-dimensional one-hot piece indicator, while the hub stores 22 global features (castling rights, en passant location, and scaled half- and full-move clocks), resulting in 34-dimensional feature vectors for all nodes. 
Edges comprise (i) all potential moves for both sides and (ii) bidirectional links between each square and the hub. 
Each edge is annotated with a 7-bit attribute indicating legality, capture, defense, promotion, side, forward/backward direction, and local vs. global connection.

We first project all node and edge features into \(\mathbb{R}^{128}\) via learnable linear layers. 
The graph is then fed through two consecutive GINEConv layers \cite{pyg-gineconv}, updating each node feature vector by aggregating its neighbors and edge attributes. 
Global average and max pooling of node features yield a 256-dimensional embedding, which is passed to the fixed-weight recurrent BPU.

\textbf{CNN-based encoder.}  
To match the parameter count of Transformer models in ChessBench [8], we tokenize FEN into a $[24, 8, 8]$ tensor comprising 22 semantic channels (12 piece types, side-to-move, castling rights, en passant, move counters, promotion indicators) and 2 spatial coordinate channels. The tensor is processed by a six-layer convolutional encoder: a two-layer convolutional stem $3\times3$ followed by six residual blocks with alternating Squeeze-and-Excitation (SE) modules and stochastic depth. 
Global average pooling yields a 256-dimensional embedding. The embedding is passed to the fixed-weight recurrent BPU, with only the input and output projections trainable. 

We evaluate three variants: GNN, CNN, and CNN enhanced with a minimax search and alpha-beta pruning during inference \cite{shannon1950chess,gundawar2024superiorcomputerchessmodel,KNUTH1975293,fuller1973analysis}, which refines move selection without increasing model capacity. The precision of the puzzle is measured under the training budgets of the games $\{10^4$, $10^5$,  $10^6\}$, benchmarked against previous results from ChessBench~\cite{lappalainen2024connectome}. Final evaluation is performed on 10,000 curated Lichess puzzles, each with full solution sequences and Elo scores.

\section{Results}
\subsection{Image Classification}
\begin{figure}[!ht]
    \centering
    \includegraphics[width=\linewidth]{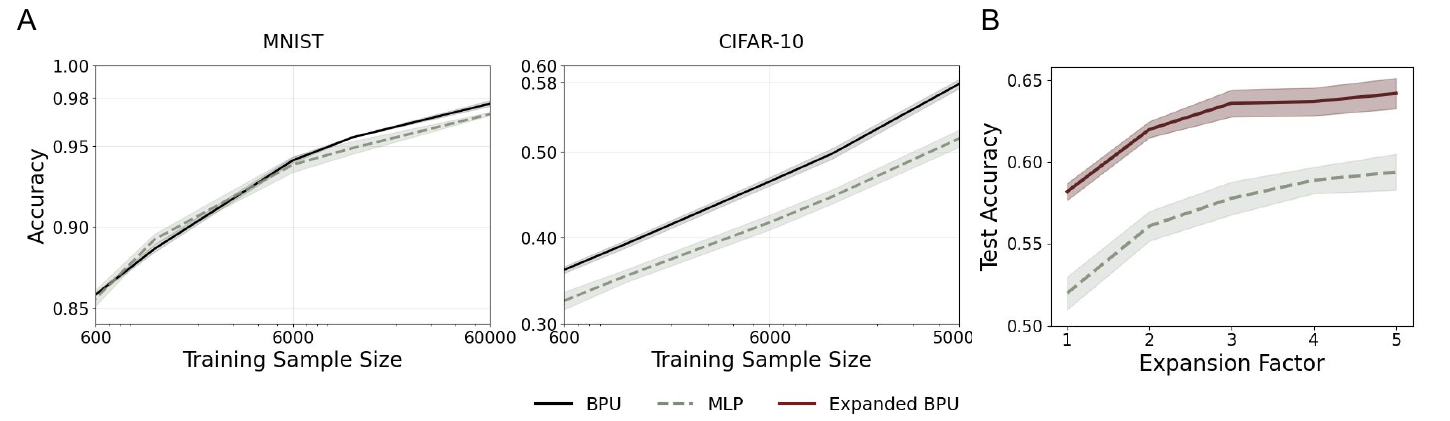}
    \caption{\textbf{(A)} Test accuracy on MNIST and CIFAR-10 for the original connectome-derived BPU.  \textbf{(B)} CIFAR-10 test accuracy as a function of expansion factor for expanded BPUs via DCSBM. Shaded bands indicate average over five runs and are compared to a size-matched 2-layer MLP baseline.} %

    \label{fig:img-cls}
\end{figure}

Figure \ref{fig:img-cls}A summarizes test accuracies on MNIST and CIFAR‑10 for two untrainable architectures: the full‑connectome BPU and a two‑layer MLP baseline with matched projection parameters. On MNIST, the full‑connectome reservoir peaks at 98$\%$ test accuracy, compared to 97$\%$ for the MLP. On the more challenging CIFAR‑10 task, the full reservoir reaches 58$\%$ while the MLP achieves 52$\%$. These performance gaps persist across small to full training set sizes.

To probe whether additional fly-like circuitry can push performance further, we expanded the connectome with the signed DCSBM generator and froze the resulting recurrent weights.  Figure~\ref{fig:img-cls}B shows that CIFAR-10 accuracy grows monotonically with expansion factor: a $2{\times}$ graph already surpasses the original BPU, and performance continues to climb, remaining consistently above the size-matched MLP baseline. Thus, scaling the biological prior yields clear benefits without any extra training of the recurrent matrix.

\begin{figure}[!ht]
    \centering
    \includegraphics[width=0.7\linewidth]
    {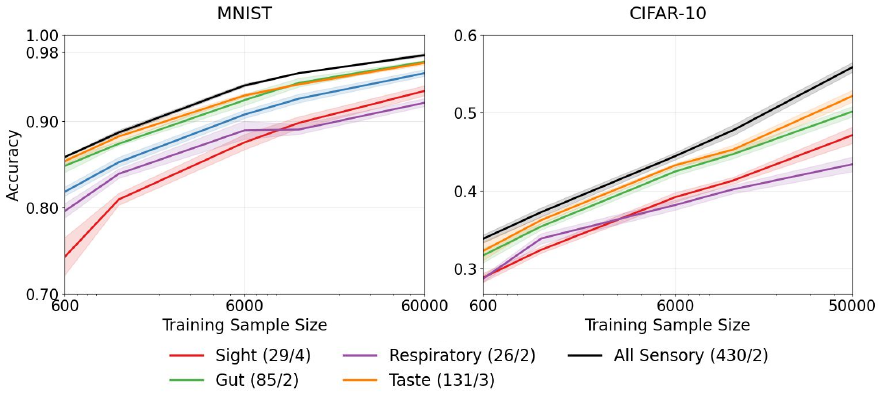}
    \caption{Test accuracy on MNIST and CIFAR-10 with modality-restricted reservoirs. Parentheses indicate (number of neurons / time steps) for each sensory subset.}
    \label{fig:img-ablation}
\end{figure}

Figure~\ref{fig:img-ablation} shows an ablation study that evaluates the contribution of different sensory modalities to image classification. Performance does not scale directly with neuron count, e.g., the respiratory group (26 neurons) outperforms the larger sight-related group (29 neurons) when trained with a small training sample size, highlighting the role of functional specificity. This may reflect evolved relevance of certain modalities, or alternatively, developmental limitations of the 6-hour-old larva\cite{winding2023connectome}, where some circuits may be immature.

\subsection{Chess Puzzle Solving}
\begin{figure}[!ht]
    \centering
    \includegraphics[width=0.58\linewidth]{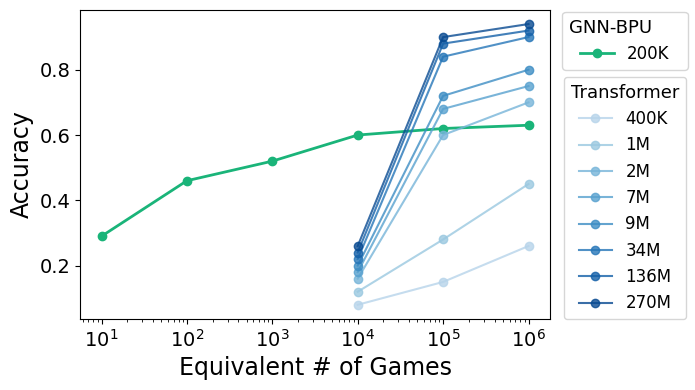}
    \caption{Puzzle‐solving accuracy (\%) with GNN--BPU model and ChessBench reference models of multiple sizes. Despite having only 232$,$912 trainable parameters, the GNN--BPU converges even with small dataset size and achieves competitive or superior accuracy to the baselines.}
    \label{fig:Chess_gnn}
\end{figure}
We evaluated our GNN--BPU model, which contains only 232,912 trainable parameters, on the ChessBench dataset \cite{ruoss2024amortized} under multiple training budgets. As illustrated in Figure~\ref{fig:Chess_gnn}, the model attains accuracies of 59\%, 61\%, and 63\% when trained on $10^{4}$, $10^{5}$, and $10^{6}$ games, respectively. Remarkably, the GNN–BPU performs strongly with even smaller datasets. It also consistently surpasses the smallest reference model from Ruoss et al. \cite{ruoss2024amortized}, despite that baseline still having more parameters—and remains competitive with substantially larger models. These results underscore the effectiveness of our biologically inspired reservoir architecture for data-efficient strategic reasoning.

\begin{figure}[!ht]
    \centering
    \includegraphics[width=1.04\linewidth]{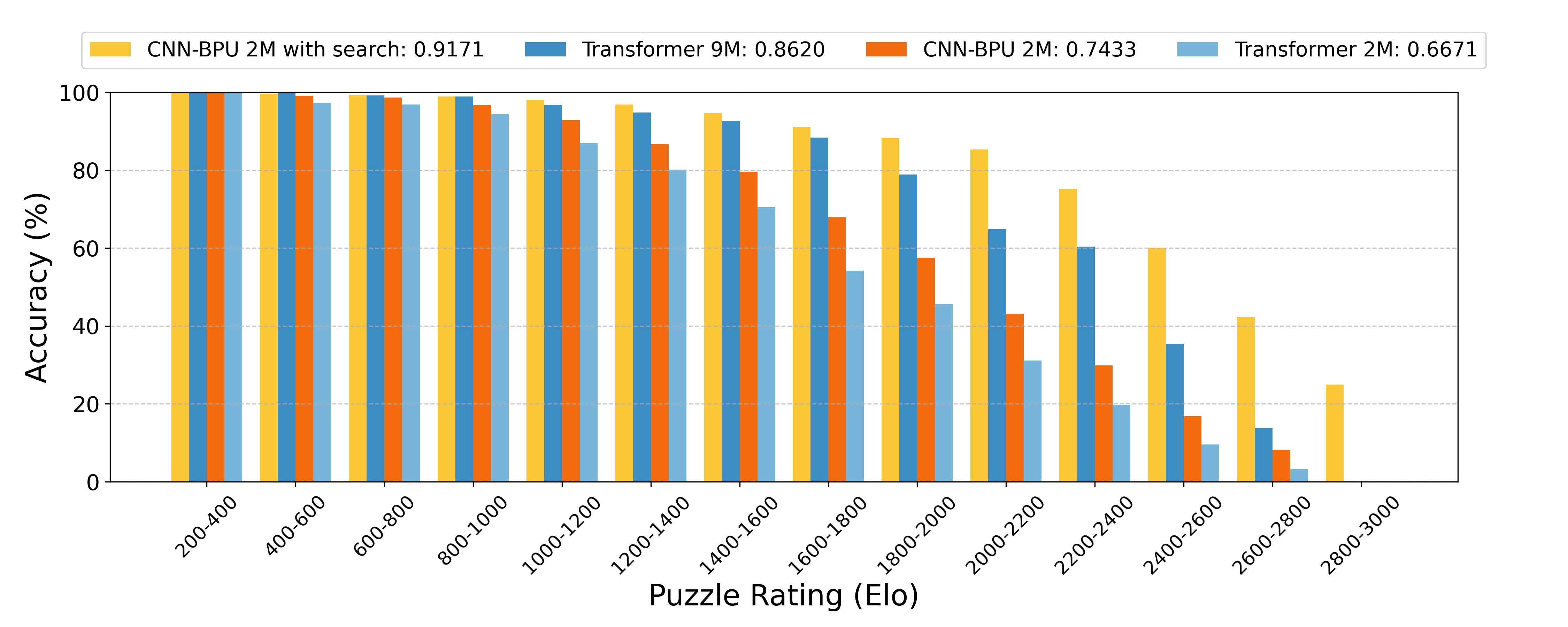} 
    \caption{Bars show the percentage of puzzles solved correctly within each Elo bin. The legend indicates model type, parameter count, and overall accuracy. At equal scale ($\sim$2M), CNN--BPU outperforms the Transformer baseline. With search, CNN--BPU surpasses even a 9M-parameter Transformer.} 
    \label{fig:puzzles}
\end{figure}
To further assess scalability, we investigate whether the BPU remains competitive at the same parameter scale as Transformer baselines. As shown in Figure~\ref{fig:puzzles}, the CNN--BPU model with $\sim$2M parameters outperforms the Transformer of equivalent size. When equipped with a minimax search of depth~6 and alpha--beta pruning at inference, CNN--BPU achieves 91.71\% puzzle accuracy, surpassing even the 9M-parameter Transformer baseline.

To ensure a fair comparison, we reimplement the 2M-parameter Transformer using the open-source code from~\cite{ruoss2024amortized}, and directly evaluate the official pretrained 9M-parameter checkpoint. All models are assessed using puzzle accuracy across Elo bins, as shown in Figure~\ref{fig:puzzles}.

\section{Discussion}
Our results demonstrate that the complete \textit{Drosophila} larval connectome,  without any structural modification, can serve as an  efficent neural substrate for complex tasks such as image recognition and chess puzzle solving. This suggests that even circuits evolved for simpler behaviors possess a significant latent computational capacity.

To clearly isolate this intrinsic capacity, we intentionally avoided any structural rewiring or synaptic tuning. While this approach highlights the connectome's inherent capabilities, performance could likely be enhanced. Future work could explore refining the connectome with task-specific adaptations, such as structure-aware rescaling or constrained plasticity mechanisms \cite{pedigo2023generative,geroldinger2021arithmeticmonoidsideals}, without losing its biological inductive priors. 

Another important direction for future research is understanding how different parts of the connectome contribute to task performance. Our ablation studies focused on sensory neuron types, but functional specialization may depend on richer circuit motifs, such as feedback loops \cite{vishwanathan2024predicting}, recurrent clusters, or region-specific pathways that cannot be captured by simple type-based removal. Elucidating the causal roles of these substructures remains an important open question.

Finally, the connectome used here is from a larva only a few hours post-hatching. While it provides a complete, compact testbed, its behavioral repertoire is limited. As more comprehensive adult or cross-species connectomes become available, it will be crucial to evaluate whether the same principles scale to larger and more cognitively capable brains, such as the adult \textit{Drosophila}~\cite{shiu2024drosophila}, and eventually the human connectome. The ultimate goal—though ambitious—is clear: leveraging detailed connectomic data, starting from the simplest complete brain structures, to build increasingly intelligent and capable AI.

\subsubsection{Acknowledgments}
We acknowledge support from the National Science Foundation (Grant No.,20-540). We also thank Yuxin Bai for insightful feedback and suggestions.
\bibliographystyle{unsrt}
\bibliography{bib}

\begin{thebibliography}{10}

\bibitem{winding2023connectome}
Michael Winding, Benjamin~D Pedigo, Christopher~L Barnes, Heather~G Patsolic, Youngser Park, Tom Kazimiers, Akira Fushiki, Ingrid~V Andrade, Avinash Khandelwal, Javier Valdes-Aleman, et~al.
\newblock The connectome of an insect brain.
\newblock {\em Science}, 379(6636):eadd9330, 2023.

\bibitem{frankle2019lotterytickethypothesisfinding}
Jonathan Frankle and Michael Carbin.
\newblock The lottery ticket hypothesis: Finding sparse, trainable neural networks, 2019.

\bibitem{pmlr-v119-hasani20a}
Ramin Hasani, Mathias Lechner, Alexander Amini, Daniela Rus, and Radu Grosu.
\newblock A natural lottery ticket winner: Reinforcement learning with ordinary neural circuits.
\newblock In Hal~Daumé III and Aarti Singh, editors, {\em Proceedings of the 37th International Conference on Machine Learning}, volume 119 of {\em Proceedings of Machine Learning Research}, pages 4082--4093. PMLR, 13--18 Jul 2020.

\bibitem{lappalainen2024connectome}
Juho~K Lappalainen, Fabian~D Tschopp, Sai Prakhya, et~al.
\newblock Connectome-constrained networks predict neural activity across the fly visual system.
\newblock {\em Nature}, 634:1132--1140, 2024.

\bibitem{wang2023neuroinspired}
Lei Wang, Xiaohui Zhang, Qian Li, et~al.
\newblock Incorporating neuro-inspired adaptability for continual learning in artificial intelligence.
\newblock {\em Nature Machine Intelligence}, 5:1356--1368, 2023.

\bibitem{liang2021fruitflylearnword}
Yuchen Liang, Chaitanya~K. Ryali, Benjamin Hoover, Leopold Grinberg, Saket Navlakha, Mohammed~J. Zaki, and Dmitry Krotov.
\newblock Can a fruit fly learn word embeddings?, 2021.

\bibitem{zheng2018complete}
Zhihao Zheng, Jason~S Lauritzen, Eric Perlman, Craig~G Robinson, Matthew Nichols, Daniel Milkie, Oriol Torrens, Jackson Price, Conrad~B Fisher, Nick Sharifi, Sarah~A Calle-Schuler, Lenka Kmecova, Iman~J Ali, Benjamin Karsh, Emily~T Trautman, John~A Bogovic, Philipp Hanslovsky, Gregory S X~E Jefferis, Michael Kazhdan, Khaled Khairy, Stephan Saalfeld, Richard~D Fetter, and Davi~D Bock.
\newblock A complete electron microscopy volume of the brain of adult drosophila melanogaster.
\newblock {\em Cell}, 174(3):730--743.e22, 2018.

\bibitem{ruoss2024amortized}
Anian Ruoss, Gr\'egoire Del\'etang, Sourabh Medapati, Jordi Grau-Moya, Li~Kevin Wenliang, Elliot Catt, John Reid, Cannada~A. Lewis, Joel Veness, and Tim Genewein.
\newblock Amortized planning with large-scale transformers: A case study on chess.
\newblock {\em arXiv preprint arXiv:2402.04494}, 2024.

\bibitem{wang2024we}
Qingyang Wang, Albert Cardona, Marta Zlatic, Joshua~T Vogelstein, and Carey~E Priebe.
\newblock Why do we have so many excitatory neurons?
\newblock {\em bioRxiv}, pages 2024--09, 2024.

\bibitem{karrer2011stochastic}
Brian Karrer and Mark~EJ Newman.
\newblock Stochastic blockmodels and community structure in networks.
\newblock {\em Physical Review E—Statistical, Nonlinear, and Soft Matter Physics}, 83(1):016107, 2011.

\bibitem{gao2018community}
Chao Gao, Zongming Ma, Anderson~Y Zhang, and Harrison~H Zhou.
\newblock Community detection in degree-corrected block models.
\newblock {\em The Annals of Statistics}, 2018.

\bibitem{carlini2023playing}
Nicholas Carlini.
\newblock Playing chess with large language models, 2023.

\bibitem{edwards1994standard}
SJ~Edwards, SD~Forsyth, J~Stanback, and A~Saremba.
\newblock Standard portable game notation specification and implementation guide. 1994.
\newblock {\em URL https://ia902908. us. archive. org/26/items/pgn-standard-1994-03-12/PGN standard}, pages 03--12, 1994.

\bibitem{pyg-gineconv}
{PyTorch Geometric Development Team}.
\newblock {torch\_geometric.nn.conv.GINEConv} — pytorch geometric 2.5.1 documentation, 2024.

\bibitem{shannon1950chess}
Claude~E. Shannon.
\newblock Programming a computer for playing chess.
\newblock {\em Philosophical Magazine}, 41:256--275, 1950.
\newblock Introduced the Minimax algorithm in chess.

\bibitem{gundawar2024superiorcomputerchessmodel}
Atharva Gundawar, Yuchao Li, and Dimitri Bertsekas.
\newblock Superior computer chess with model predictive control, reinforcement learning, and rollout, 2024.

\bibitem{KNUTH1975293}
Donald~E. Knuth and Ronald~W. Moore.
\newblock An analysis of alpha-beta pruning.
\newblock {\em Artificial Intelligence}, 6(4):293--326, 1975.

\bibitem{fuller1973analysis}
Samuel~H Fuller, John~G Gaschnig, JJ~Gillogly, et~al.
\newblock {\em Analysis of the alpha-beta pruning algorithm}.
\newblock Department of Computer Science, Carnegie-Mellon University, 1973.

\bibitem{pedigo2023generative}
Benjamin~D Pedigo, Mike Powell, Eric~W Bridgeford, Michael Winding, Carey~E Priebe, and Joshua~T Vogelstein.
\newblock Generative network modeling reveals quantitative definitions of bilateral symmetry exhibited by a whole insect brain connectome.
\newblock {\em Elife}, 12:e83739, 2023.

\bibitem{geroldinger2021arithmeticmonoidsideals}
Alfred Geroldinger and M.~Azeem Khadam.
\newblock On the arithmetic of monoids of ideals, 2021.

\bibitem{vishwanathan2024predicting}
A.~Vishwanathan, A.~Sood, J.~Wu, et~al.
\newblock Predicting modular functions and neural coding of behavior from a synaptic wiring diagram.
\newblock {\em Nature Neuroscience}, 27:2443--2454, 2024.

\bibitem{shiu2024drosophila}
P.~K. Shiu, G.~R. Sterne, N.~Spiller, et~al.
\newblock A drosophila computational brain model reveals sensorimotor processing.
\newblock {\em Nature}, 634:210--219, 2024.

\end{thebibliography}
%


\end{document}